\documentclass[conference]{IEEEtran}
\IEEEoverridecommandlockouts

\usepackage{spconf}
\usepackage{cite}
\usepackage{amsmath,amssymb,amsfonts}
\usepackage{algorithmic}
\usepackage{graphicx}
\usepackage{textcomp}
\usepackage{xcolor}
\usepackage{pifont}
\usepackage{url}

\def\BibTeX{{\rm B\kern-.05em{\sc i\kern-.025em b}\kern-.08em
    T\kern-.1667em\lower.7ex\hbox{E}\kern-.125emX}}

\begin{document}

\title{Lightweight Wasserstein Audio-Visual Model for Unified Speech Enhancement and Separation}


\maketitle

\begin{abstract}
Speech Enhancement (SE) and Speech Separation (SS) have traditionally been treated as distinct tasks in speech processing. However, real-world audio often involves both background noise and overlapping speakers, motivating the need for a unified solution. While recent approaches have attempted to integrate SE and SS within multi-stage architectures, these approaches typically involve complex, parameter-heavy models and rely on supervised training, limiting scalability and generalization. In this work, we propose \textit{UniVoiceLite}, a lightweight and unsupervised audio-visual framework that unifies SE and SS within a single model. UniVoiceLite leverages lip motion and facial identity cues to guide speech extraction and employs Wasserstein distance regularization to stabilize the latent space without requiring paired noisy-clean data. Experimental results demonstrate that UniVoiceLite achieves strong performance in both noisy and multi-speaker scenarios, combining efficiency with robust generalization. The source code is available at \url{https://github.com/jisoo-o/UniVoiceLite}.
\end{abstract}

\section{Introduction}

Speech processing has witnessed remarkable advancements in recent years \cite{xu2014regression,wang2018supervised}, driven by the growing demand for robust speech enhancement (SE) and speech separation (SS) in various applications such as telecommunication, hearing aids, and automatic speech recognition. SE aims to extract a target speaker's voice from noisy environments, improving intelligibility and perceptual quality, while SS focuses on disentangling overlapping voices in a mixture to enable individual speaker identification and transcription \cite{erdogan2015phase,hershey2016deep}. However, real-world environments rarely contain isolated sounds; instead, they often involve a mixture of background noise and multiple speakers. This has led to growing interest in unified frameworks that can jointly handle both tasks.

\begin{figure}[t]
  \centering
   \includegraphics[width=1\linewidth]{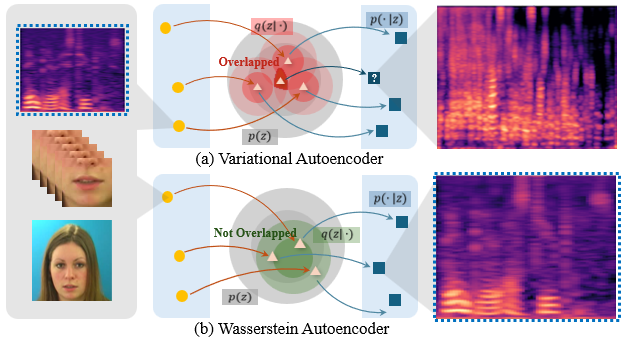}
   \vspace{-5mm}
   \caption{\textbf{Advantages of the Proposed WAE for Integrated Speech Enhancement and Separation Tasks.} (a) Traditional VAE exhibits significant overlap in the latent space due to many-to-one mapping, leading to blurry and less precise reconstructions. (b) The proposed WAE reduces this overlap and produces reconstructions that more accurately  resemble the clean input (highlighted with blue dashed boxes), making it well-suited for integrated speech enhancement and separation tasks.}
   \label{fig:vae,wae}
   \vspace{-5mm}
\end{figure}

To address this issue, recent studies have proposed multi-stage architectures that incorporate denoising modules into the separation pipeline. For instance, MUSE~\cite{saijo2023single} integrates a denoising front-end with a target speaker extraction module, enabling joint enhancement and separation within a single system. While effective, such unified approaches often rely on complex, parameter-heavy architectures and require supervised training with noisy-clean speech pairs. These constraints raise concerns regarding computational efficiency, scalability, and the ability to generalize to unseen conditions.

To ovecome the aforementioned challenges, we propose \textit{UniVoiceLite}, a simple yet effective unsupervised audio-visual framework that unifies SE and SS within a lightweight model architecture. Our method eliminates the need for paired noisy-clean data and, \textit{more importantly, directly models clean speech distributions to achieve robust generalization across diverse acoustic environments, while remaining speaker-independent.} Table~\ref{tab:novelty_comparison} highlights the uniqueness of our model design by comparing existing approaches in terms of three key aspects: unified handling of SE and SS, unsupervised learning, and lightweight architecture (fewer than 3 million parameters).

\noindent\textbf{Advantages 1.}  
UniVoiceLite is designed to be both visually guided and lightweight. By incorporating lip motion and facial identity features as visual priors, the model improves speech intelligibility and target speaker discrimination, especially in noisy and overlapping conditions. Unlike conventional architectures that rely on large convolutional backbones or multi-stage pipelines, UniVoiceLite uses shallow linear layers for each modality (audio, lip, identity), followed by a shared latent space and a unified decoder. This compact design enables the model to operate effectively with only \texttt{2.3M} parameters, significantly fewer than existing audio-visual separation models such as VisualVoice~\cite{gabbay2017visual} (\texttt{77.75M}), making UniVoiceLite suitable for deployment in resource-constrained settings.

\noindent\textbf{Advantages 2.}  
Existing speech enhancement methods based on Variational Autoencoder (VAE) \cite{VAE} often suffer from posterior collapse, a phenomenon where latent variables collapse to the prior distribution, leading to degraded speech quality \cite{AV-VAE,leglaive2018variance,fang2021variational}. This issue arises due to the strong regularization imposed by the Kullback-Leibler (KL) divergence, which constrains the model’s ability to learn meaningful speech structures. In order to alleviate this, we leverage the Wasserstein Autoencoder (WAE) \cite{tolstikhin2017wasserstein}, which replaces the KL regularization with the Wasserstein distance, leading to a more stable and structured latent space. As a result, our model effectively mitigates posterior collapse and generates cleaner, higher-quality speech, as shown in Fig.\ref{fig:vae,wae}.

\begin{figure*}[!th]
\centering
\includegraphics[width=0.9\linewidth]{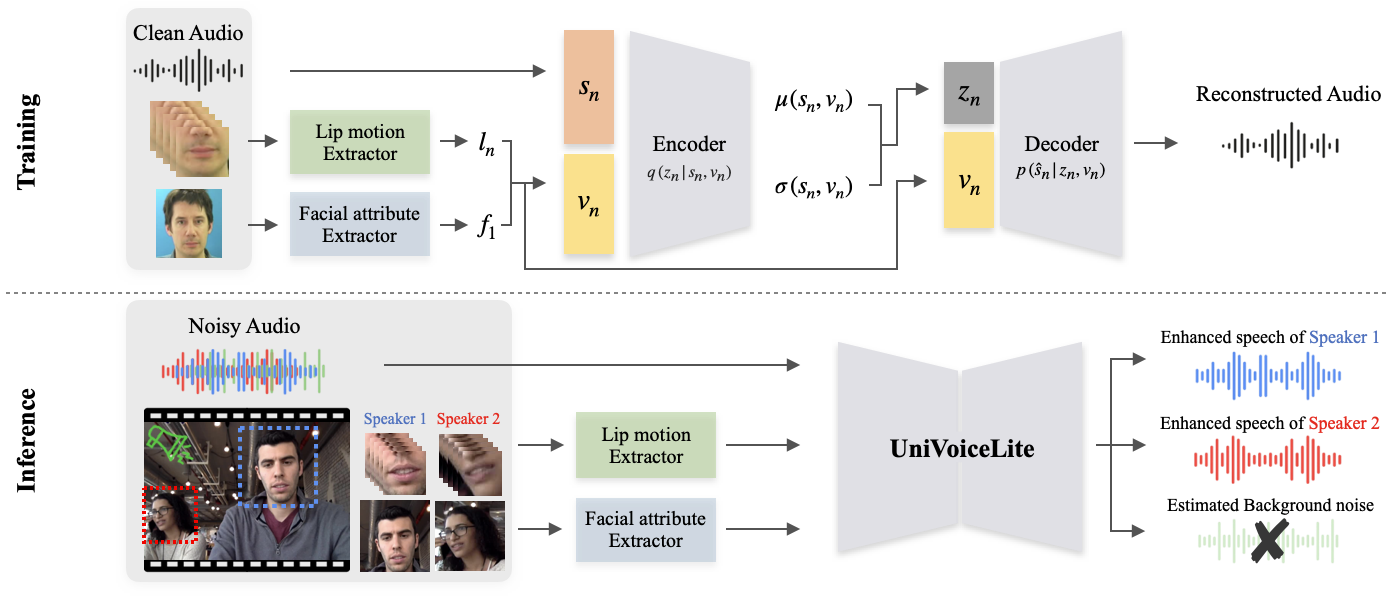}
\vspace{-5mm}
\caption{\textbf{Illustration of the Proposed UniVoiceLite Pipeline.} During \textit{training}, the model takes visual features $v_n$, including lip motion and facial attributes, along with clean speech features ${s}_n$ as inputs to the encoder, which maps them to a latent space. The decoder then reconstructs the speech signal using the visual prior $v_n$. During \textit{inference}, the model processes noisy audio and speaker-specific visual information from multiple speakers. UniVoiceLite then separates the speech components, generating enhanced speech for each speaker while effectively filtering out background noise.}
\label{fig:network}
\vspace{-3mm}
\end{figure*}

The main contributions can be summarized as follows.

\begin{itemize}
\item We propose \textit{UniVoiceLite}, a lightweight, unsupervised framework that unifies speech enhancement and speech separation, leveraging visual information for improved speech processing.
\item We incorporate the Wasserstein distance to regularize the latent space, ensuring a more stable and structured representation, enhancing the quality of the learned representations.
\item We show the effectiveness of our approach through extensive experiments in noisy, multi-talker environments, achieving competitive improvements in speech intelligibility and clarity, both quantitatively and qualitatively. 
\end{itemize}

\section{Related Works}

\textbf{Audio Features.} In single-microphone speech processing, magnitude spectrograms are widely adopted due to their effectiveness in capturing the energy distribution across frequency and time. However, they inherently discard phase information, which is critical for natural-sounding speech reconstruction. To overcome this, several studies have explored richer spectral representations. These approaches include incorporating phase information \cite{afouras2018conversation, afouras2019my} from complex-valued Fourier Transform coefficients, using the real and imaginary components of the complex spectrogram \cite{ephrat2018looking, ideli2019audio, inan122019evaluating}, or even employing the raw waveform directly as input features \cite{ideli2019visually} for speech enhancement. These alternative representations preserve signal details that the magnitude spectrogram alone may overlook, improving performance in speech processing. More recently, self-supervised learning on audio such as wav2vec 2.0~\cite{baevski2020wav2vec} has also been explored for speech enhancement and recognition, suggesting a shift toward learning more abstract and robust representations from raw audio. 

\begin{table}[t]
\centering
\setlength{\tabcolsep}{5pt}
\caption{\textbf{Comparison of model design novelty.} UniVoiceLite is the only method that is unified, unsupervised, and lightweight.}
\label{tab:novelty_comparison}
\vspace{1mm}
\begin{tabular}{lccc}
\hline
\textbf{Method} & \textbf{Unified Model} & \textbf{Unsupervised} & \textbf{Lightweight} \\
\hline

MUSE~\cite{saijo2023single}      & O & × & × \\
RVAE~\cite{RVAE}                 & × & O & O\\
AV-VAE~\cite{AV-VAE}             & × & × & O\\
MP-SENet~\cite{lu2023mp}         & × & × & O\\
CMGAN~\cite{cao2022cmgan}        & × & × & O \\
HiFi-GAN~\cite{kong2020hifi}     & × & × & O \\
MossFormer2~\cite{zhao2024mossformer2} & O & × & × \\
VisualVoice~\cite{gabbay2017visual}   & O & × & × \\
\textbf{UniVoiceLite (Ours)}     & O & O & O \\

\hline
\vspace{-5mm}
\end{tabular}
\end{table}

\noindent\textbf{Visual Features.} Visual information such as lip motion and facial attributes have been shown to significantly improve speech enhancement and separation in noisy environments, where audio-only methods struggle. Visual information is invariant to acoustic noise and complements audio when signals are corrupted or overlapping. Several audio-visual models~\cite{ephrat2018looking, afouras2018deep, gabbay2017visual} demonstrate the effectiveness of leveraging lip movements to localize and enhance the target speaker. Static identity embeddings, such as those from facial recognition models~\cite{parkhi2015deep, schroff2015facenet}, also serve as priors to guide speech extraction in multi-speaker settings~\cite{afouras2020self}. For instance, AV-HuBERT~\cite{shi2022learning} learns multimodal representations using lip region features combined with contrastive and clustering objectives, which has shown strong performance in speech tasks under weak supervision. Furthermore, temporal modeling using transformers~\cite{kalkhorani2023time} or LSTMs~\cite{jain2024lstmse} allows the system to capture both dynamic articulations and speaker consistency across time.

\noindent\textbf{Variational Autoencoder.}
In recent speech enhancement approaches, VAEs \cite{VAE} have gained attention for their ability to perform unsupervised learning, eliminating the need for labeled clean speech data. In addition, VAEs learn the latent space of speech signals, facilitating effective noise separation and improving reconstruction performance. AV-VAE \cite{AV-VAE} utilized a VAE to combine visual and auditory information for speech reconstruction, achieving better performance than single-modality approaches. Meanwhile, Similarly, \cite{RVAE} employed a Recurrent VAE to effectively model temporal dependencies, resulting in more robust speech reconstruction. 

In contrast to these models, our approach leverages the WAE to mitigate posterior collapse, ensuring a more structured latent space. Additionally, by integrating visual features, our model enhances speech clarity and robustness across diverse acoustic environments.

\noindent\textbf{Wasserstein Autoencoder.}
WAE \cite{tolstikhin2017wasserstein} improved upon VAEs \cite{VAE} by addressing posterior collapse and reducing latent space overlap, both of which can degrade speech reconstruction. Unlike VAEs, which impose a KL divergence constraint, WAE regularizes the latent space using the Wasserstein distance, enforcing a more structured and meaningful latent representation while preventing latent space collapse.

By incorporating WAE into UniVoiceLite, we ensure a more stable latent representation, reducing artifacts and improving speech intelligibility. As illustrated in Fig.\ref{fig:vae,wae}, WAE effectively mitigates the many-to-one mapping issue observed in VAEs, leading to clearer and more natural speech reconstruction.

\noindent\textbf{Unified Speech Enhancement and Separation.}
Traditional SE and SS tasks have typically been addressed with separate models~\cite{wang2018supervised}. However, real-world applications often demand the simultaneous handling of both. To meet this need, recent models~\cite{saijo2023single,gabbay2017visual} attempted to integrate enhancement and separation into a single framework. These models, however, typically rely on supervised training with large paired datasets and complex architectures involving multiple encoder-decoder stacks or multi-stage refinement modules. For example, the method proposed in~\cite{saijo2023single} introduced a denoising module prior to a target speaker extractor, while the approach in~\cite{gabbay2017visual} combined lip motion and audio encoders with a fusion network. Despite their performance, such models are computationally expensive and require strong supervision.

In contrast, our model offers a unified solution that is lightweight, unsupervised, and effective across both SE and SS tasks.

\section{Unified Wasserstein Auto-Encoder}
\label{sec:method}

The input speech feature $\bold{s}_n = \left(\left|s^0_{n}\right|^2,\left|s^1_{n}\right|^2, \ldots, \left|s_n^{F-1}\right|^2\right)$ is defined as the power spectrum of the complex-valued STFT coefficients $s^f_n$, where $n$ and $f$ denote the time frame and frequency bin, respectively. This representation captures the time-frequency structure of speech signals essential for modeling both content and noise characteristics.

Each audio frame $\bold{s}_n$ is paired with a synchronized visual input $\bold{v}_n = (l_n, f_1)$, where $l_n$ represents dynamic lip motion features extracted from video frames around time $n$, and $f_1$ denotes static facial identity features obtained from the first frame. These visual cues provide complementary information useful for separating and enhancing speech under challenging acoustic conditions.

The dataset comprises a sequence of paired audio-visual frames $(\bold{s}_n, \bold{v}_n)$, serving as the input for our unsupervised learning framework. Unlike conventional supervised approaches that require paired clean and noisy speech, our method leverages only clean speech and aligned visual inputs, enabling scalable training without reliance on curated ground-truth references.

In this section, we propose \textbf{UniVoiceLite}, a unified and lightweight audio-visual Wasserstein auto-encoder that jointly performs speech enhancement and separation by integrating visual guidance into both latent representation learning and speech reconstruction. The model encodes speech-relevant information into a latent variable $\bold{z}_n$, which is regularized via a visually conditioned prior. This structure prevents posterior collapse and promotes robust speech generation across both single- and multi-speaker scenarios.

\begin{table*}[t]
\caption{\textbf{\textcolor{black}{Speech Enhancement (SE)}} Comparison Using SDR, STOI, DNSMOS-s, and DNSMOS-o metrics.
The \textbf{best} and \underline{second best} results were highlighted.}
\label{tab:noisy}
\vspace{1mm}
\centering
\footnotesize{
\begin{tabular}{ccccc|cccc}
\hline
              & SDR\(\uparrow\) & STOI\(\uparrow\) & DNSMOS-s\(\uparrow\) & DNSMOS-o\(\uparrow\) & SDR\(\uparrow\) & STOI\(\uparrow\) & DNSMOS-s\(\uparrow\) & DNSMOS-o\(\uparrow\) \\ \hline
Noise         & \multicolumn{4}{c|}{Station}          & \multicolumn{4}{c}{Kitchen}           \\ \hline

RVAE {\scriptsize \textcolor{gray}{2020}}          & -7.28±5.11 & 0.53±0.12 & 1.18±0.01 & 1.08±0.01  
              & \underline{-1.21±2.79} & \underline{0.64±0.12} & 1.62±0.38 & 1.22±0.12  \\
AV-VAE {\scriptsize \textcolor{gray}{2021}}        & 0.48±2.13 & 0.30±0.03 & 1.64±0.14 & 1.39±0.07 
              & -2.01±2.81 & 0.29±0.01 & 1.84±0.26 & 1.51±0.17  \\
MP-SENet {\scriptsize \textcolor{gray}{2023}}       & -14.54±0.72 & 0.15±0.02 & \textbf{3.37±0.14} & \textbf{3.10±0.17} 
              & -14.63±0.06 & 0.18±0.01 & \textbf{3.45±0.03} & \textbf{3.18±0.04}  \\
CMGAN {\scriptsize \textcolor{gray}{2022}}       & -13.73±0.23 & 0.15±0.02 & \underline{3.20±0.14} & \underline{2.90±0.17} 
              & -14.30±0.07 & 0.18±0.01 & \underline{3.23±0.17} & \underline{2.90±0.19} \\
HiFi-GAN {\scriptsize \textcolor{gray}{2020}}       & \underline{0.55±6.95} & \underline{0.60±0.14} & 2.51±1.17 & 1.76±0.64 
              & 0.13±7.25 & 0.61±0.15 & 3.13±0.06 & 2.13±0.30  \\
\textbf{Ours} & \textbf{2.11±6.47} & \textbf{0.67±0.12} & 2.40±1.06 & 1.53±0.42  
              & \textbf{3.10±6.03} & \textbf{0.69±0.13} & 3.05±0.06 & 1.78±0.14  \\\hline
Noise         & \multicolumn{4}{c|}{Metro}            & \multicolumn{4}{c}{Cafeteria}         \\ \hline
RVAE {\scriptsize \textcolor{gray}{2020}}          & -3.40±1.65 & \underline{0.65±0.11} & 1.59±0.23 & 1.33±0.12  
              & -7.81±5.94 & \underline{0.49±0.12} & 1.16±0.04 & 1.07±0.02  \\
AV-VAE {\scriptsize \textcolor{gray}{2021}}      & -1.72±2.99 & 0.25±0.05 & 1.71±0.08 & 1.44±0.03
              & \underline{0.16±1.78} & 0.31±0.05 & 1.75±0.12 & 1.45±0.05  \\
MP-SENet {\scriptsize \textcolor{gray}{2023}}      & -13.91±0.27 & 0.16±0.02 & \textbf{3.46±0.08} & \textbf{3.19±0.09} 
              & -14.60±0.17 & 0.18±0.04 & \textbf{3.32±0.28} & \textbf{3.03±0.34}  \\
CMGAN {\scriptsize \textcolor{gray}{2022}}      & -13.79±0.30 & 0.19±0.01 & \underline{3.20±0.11} & \underline{2.90±0.11} 
              & -14.22±0.12 & 0.17±0.01 & \underline{3.06±0.51} & \underline{2.75±0.55}  \\
HiFi-GAN {\scriptsize \textcolor{gray}{2020}}      & \underline{0.19±7.19} & 0.55±0.16 & 2.11±0.92 & 1.51±0.49 
              & 0.09±7.27 & 0.45±0.16 & 2.11±0.92 & 1.51±0.49  \\
\textbf{Ours} & \textbf{1.01±6.89} & \textbf{0.74±0.12} & 2.55±0.30 & 1.68±0.24  
              & \textbf{1.17±6.64} & \textbf{0.52±0.15} & 1.85±0.96 & 1.32±0.33  \\\hline
\end{tabular}%
}
\vspace{-5mm}
\end{table*}

\begin{figure}[t]
  \centering
   \includegraphics[width=1.0\linewidth]{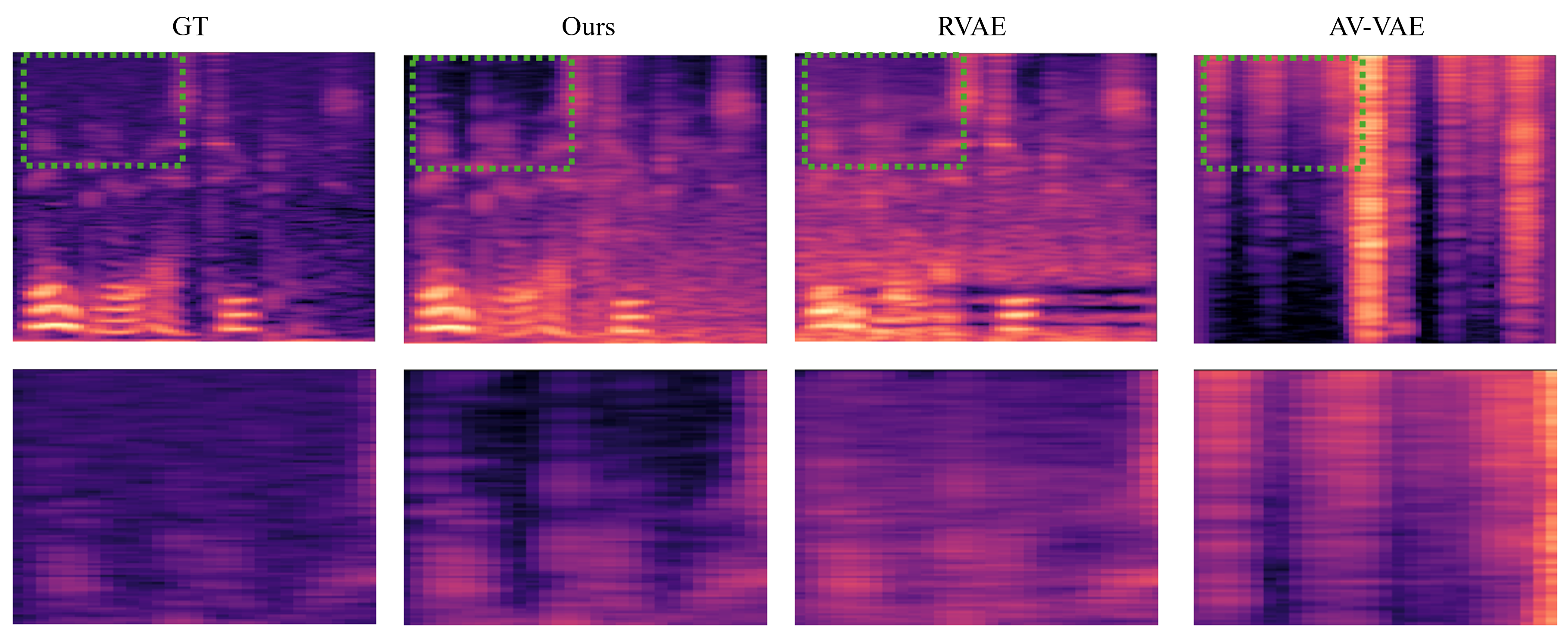}
   \vspace{-8mm}
   \caption{\textbf{\textcolor{black}{Speech Enhancement (SE)}} Comparison Using \textbf{Mel-Spectrogram}. Top: Spectrograms of ground truth, ours, RVAE, and AV-VAE. Bottom: Zoomed-in highlighted regions.}
   \label{fig:melspectrogram}
   \vspace{-4mm}
\end{figure}

\noindent\textbf{Latent Space Representation.}
To model the underlying structure of clean speech, we introduce a latent variable $\bold{z}_n$ that captures speech-relevant information conditioned on both audio and visual inputs. Given the paired features $(\bold{s}_n, \bold{v}_n)$, the encoder approximates the posterior distribution as:
\begin{equation}
\vspace{-2mm}
q(\bold{z}_n \mid \bold{s}_n, \bold{v}_n; \psi) = \mathcal{N}\left(\boldsymbol{\mu}_n, \operatorname{diag}(\boldsymbol{\sigma}_n^2)\right),
\end{equation}
where $\boldsymbol{\mu}_n$ and $\boldsymbol{\sigma}_n$ are the mean and standard deviation predicted by the encoder network, parameterized by $\psi$.

During training, we apply the reparameterization trick to sample $\bold{z}_n$ as:
\vspace{-3mm}
\begin{equation}
\vspace{-3mm}
\bold{z}_n = \boldsymbol{\mu}_n + \boldsymbol{\sigma}_n \odot \boldsymbol{\epsilon}, \quad \boldsymbol{\epsilon} \sim \mathcal{N}(\mathbf{0}, \mathbf{I}),
\end{equation}
which enables backpropagation via the sampling operation.

To regularize the latent space, we define a visually conditioned prior distribution:
\begin{equation}
\vspace{-1mm}
p(\bold{z}_n \mid \bold{v}_n; \gamma) = \mathcal{N}\left(\boldsymbol{\mu}_n^{\text{prior}}, \operatorname{diag}(\boldsymbol{\sigma}_n^{\text{prior}})\right),
\end{equation}
where $\boldsymbol{\mu}_n^{\text{prior}}$ and $\boldsymbol{\sigma}_n^{\text{prior}}$ are the mean and standard deviation vectors predicted from the visual input $\bold{v}_n$ via a neural network parameterized by $\gamma$. 

\noindent\textbf{Optimization Objective.}
The model optimizes a loss function that maintains a balance between accurate speech reconstruction and effective regularization of the latent space:
\begin{align} \label{eq:objective}
    \mathcal{L}(\bold{s}, \bold{v}; \theta, \psi) &=
    \underbrace{\mathbb{E}_{q(\bold{z}|\bold{s}, \bold{v};\psi)}[\ln p(\bold{s}|\bold{z}, \bold{v};\theta)]}_{\text{Reconstruction term}} \nonumber \\
    &\quad + \lambda \cdot 
    \underbrace{\mathcal{W}(q(\bold{z}|\bold{s}, \bold{v};\psi) \parallel p(\bold{z} | \bold{v};\gamma))}_{\text{Regularization term}},
\end{align}
where the hyperparameter $\lambda$ controls the trade-off between two objectives.

The \textit{reconstruction term} encourages the decoder (parameterized by $\theta$) to generate clean speech $\bold{s}$ that closely matches the ground truth, by maximizing the log-likelihood conditioned on the latent variable $\bold{z}$ and visual features $\bold{v}$.  

The \textit{regularization term} enforce consistency in the latent space by minimizing the Wasserstein distance $\mathcal{W}(\cdot \parallel \cdot)$ between the posterior distribution $q(\bold{z} | \bold{s}, \bold{v}; \psi)$ and the prior distribution $p(\bold{z} | \bold{v}; \gamma)$, parameterized by $\psi$ and $\gamma$, respectively. Compared to KL divergence, the Wasserstein distance provides more stable gradients and encourages a smoother and more structured latent space. The overall training pipeline is illustrated in Fig.\ref{fig:network}.

\noindent\textbf{Speech Reconstruction.}  
The decoder reconstructs the STFT coefficients of the speech signal by conditioning on  both the latent representation and the visual information:
\begin{equation}
    s^{F}_n | (\bold{z}_n, \bold{v}_n) \sim \mathcal{N}_C\left(0, \sigma_F^2 (\bold{z}_n, \bold{v}_n)\right),
\end{equation}
where $\mathcal{N}_C(0, \sigma_f^2)$ denotes a univariate complex-valued Gaussian distribution, and the variance $\sigma_F^2 (\bold{z}_n, \bold{v}_n)$ captures the structure of STFT coefficients based on the latent variable $\bold{z}_n$ and the visual features $\bold{v}_n$.
Unlike supervised methods that require access to clean reference signals, our approach learns to reconstruct clean speech in a fully unsupervised manner. 

This design makes the model inherently more robust and adaptable to real-world conditions where clean ground truth is unavailable. Crucially, the visual modality guides the latent representation toward speech-relevant features, helping the model effectively reduce noise and recover intelligible even under severe acoustic degradation.

\subsection{Audio-Visual Speech Enhancement Tasks}
Given the latent representation $\bold{z}_n$, the decoder reconstructs the clean speech signal by conditioning on both $\bold{z}_n$ and the corresponding visual features $\bold{v}_n$. the generative process is defined as:
\begin{equation}
    p\left(\bold{\hat{s}}_n \mid \bold{z_n}, \bold{v_n}; \theta\right),
\end{equation}  
where $\bold{\hat{s}}_n$ denotes the reconstructed audio signal, and $\theta$ represents the parameters of the decoder network.

This formulation enables the model to generate speech that closely approximates the original clean signal, while effectively leveraging visual information to guide the reconstruction. The visual modality provides complementary information that helps disambiguate speech content, particularly in acoustically challenging scenarios. By jointly utilizing both audio and visual inputs, the model improves the intelligibility and perceptual quality of the enhanced speech output.

\subsection{Audio-Visual Speech Separation Tasks}
UniVoiceLite performs speech separation by utilizing speaker-specific visual features to isolate each speaker's voice from a mixed audio signal. Given the mixture $\bold{s}_n^{\text{mix}}$ containing multiple speakers, the encoder estimates a latent representation, $\bold{z}_n^{(k)}$ for each speaker $k$, by conditioning  on their visual features:
\begin{equation} 
    q\left(\bold{z}_n^{(k)} | \bold{s}_n^{\text{mix}}, \bold{v}_n^{(k)};\psi\right), 
\end{equation}
where $\bold{s}_n^{mix}$ denotes the visual features corresponding to speaker $k$. The decoder then reconstructs the clean speech for each individual speaker:
\begin{equation} 
    p\left(\bold{\hat{s}}_n^{(k)} | \bold{z}_n^{(k)}, \bold{v}_n^{(k)}; \theta\right), 
\end{equation}
where $\bold{\hat{s}}_n^{(k)}$ represents the reconstructed speech of speaker $k$. This process ensures that only the target speaker’s voice is preserved, while suppressing interfering speakers and background noise. By leveraging visual information as a speaker-specific prior, the model achieves effective separation without requiring explicit speaker identity labels.
\\

\noindent To further refine the separation quality, UniVoiceLite adopts a post-processing step based on Monte Carlo Expectation Maximization (MCEM)\cite{MCEM}, following the approach of AV-VAE\cite{AV-VAE}. MCEM iteratively estimates gain and noise parameters, enabling probabilistically optimal reconstruction of the clean speech. In particular, Wiener filtering is employed within this framework to suppress residual interference, enhancing the clarity and fidelity of the separated audio.

\begin{table}[t]
\caption{\textbf{\textcolor{black}{Speech Separation (SS)}} Comparison with \textbf{Generative Methods in 2 Speakers Scenarios} Using PESQ, SDR, STOI, DNSMOS-s, and DNSMOS-o metrics.}
\label{tab:speech_separation}
\vspace{1mm}
\centering
\footnotesize{
\setlength{\tabcolsep}{1pt}
\begin{tabular}{l|ccccc}
\hline
& PESQ$\uparrow$ & SDR$\uparrow$ & STOI$\uparrow$ & DNSMOS-s$\uparrow$ & DNSMOS-o$\uparrow$ \\ \hline
RVAE                   & 1.20 ± 0.15 & -5.28 ± 4.94 & 0.53 ± 0.14 & \textbf{2.47±0.38} & \textbf{1.87±0.43} \\
AV-VAE                 & 1.23 ± 0.07 & -3.64 ± 1.72 & 0.53 ± 0.05 & 1.53±0.13          & 1.36±0.09          \\
\textbf{Ours}          & \textbf{1.27 ± 0.08} & \textbf{1.46 ± 6.16} & \textbf{0.60 ± 0.13} & \underline{2.25±0.85} & \underline{1.76±0.50} \\ \hline
\end{tabular}}
\vspace{-5mm}
\end{table}


\begin{table}[t]
\caption{\textbf{\textcolor{black}{Speech Separation (SS)}} Comparison with \textbf{State-of-the-art SE methods in 2 Speakers Scenarios} Using SDR, STOI metrics.}
\label{tab:sota_se}
\vspace{1mm}
\centering
\footnotesize{
\setlength{\tabcolsep}{13pt}
\begin{tabular}{l|c|cc}
\hline
              & \# Params     & SDR $\uparrow$         & STOI $\uparrow$        \\ \hline
MP-SENet      & \underline{2.05M}         & -14.65±0.45            & 0.16±0.03              \\
CMGAN         & \textbf{1.83M}         & -14.51±0.51            & 0.16±0.04              \\
HiFi-GAN      & --            & -0.01±7.30             & 0.60±0.17              \\
\textbf{Ours} & 2.3M & \textbf{1.46 ± 6.16}   & \textbf{0.60 ± 0.13}   \\ \hline
\end{tabular}
}
\vspace{-5mm}
\end{table}

\section{Experiments}

\subsection{Experiment Details}
\noindent\textbf{Dataset.} We conducted experiments using the GRID corpus dataset~\cite{cooke2006audio}, which includes $34$ speakers each speaking $1,000$ structured sentences. Our unsupervised model was trained solely on the clean audio without using any paired noisy-clean data. Visual input was extracted as an $88 \times 88$ mouth region utilizing Dlib~\cite{king2009dlib} and encoded via the noise-augmented AV-HuBERT~\cite{shi2022learning}.

\noindent\textbf{Evaluation.} 
We simulated two realistic conditions. In the SE setting, we added DEMAND~\cite{thiemann2013diverse} noise (station, kitchen, metro, cafeteria) to clean utterances (from GRID test set) at SNR levels of -9, 0, and 9 dB. In the SS setting, we randomly selected two or three speakers from the test set, designating one as the target speaker. The remaining speaker(s) were treated as interfering sources and mixed at different SNR levels (-9, 0 ,9 dB) with the target’s speech, following common practice in audio-visual separation benchmarks~\cite{afouras2018conversation, ephrat2018looking}. Speaker 21 was excluded due to missing video, and corrupted data were removed. While trained on GRID, which is a lab-recorded dataset, our setup evaluates generalization to real-world noise and multi-speaker interference. We reported SDR for signal fidelity, STOI for intelligibility, and DNSMOS-SIG/OVR~\cite{reddy2021dnsmos} for no-reference perceptual quality. Since our setup reflects real-world conditions, all baseline results were re-evaluated under identical settings for fair comparison.

\begin{table}[t]
\caption{\textbf{\textcolor{black}{Speech Separation (SS)}} Comparison with \textbf{State-of-the-art SS methods in 2 Speakers Scenarios} using SDR, STOI metrics.}
\label{tab:sota_ss}
\vspace{1mm}
\centering
\footnotesize{
\setlength{\tabcolsep}{10pt}
\begin{tabular}{l|c|cc}
\hline
                  & \# Params       & SDR $\uparrow$         & STOI $\uparrow$        \\ \hline
MossFormer2       & $\sim$55M          & -14.33±0.36           & 0.22±0.04             \\
VisualVoice       & $\sim$77M          & -16.29±0.51           & 0.04±0.03             \\
AV-MossFormer2    & --             & -17.04±0.79           & 0.22±0.02             \\
AV-ConvTasNet     & $\sim$13M          & -13.99±2.36           & 0.16±0.03             \\
\textbf{Ours}     & \textbf{2.3M} & \textbf{1.46 ± 6.16}  & \textbf{0.60 ± 0.13}  \\ \hline
\end{tabular}
}
\vspace{-5mm}
\end{table}
\begin{figure}[t]
  \centering
   \includegraphics[width=1\linewidth]{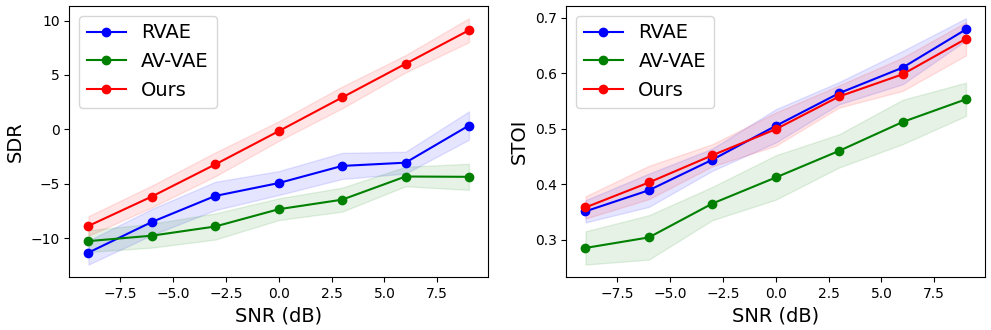}
      \vspace{-7mm}
   \caption{\textbf{\textcolor{black}{Speech Separation (SS)}} Comparison in \textbf{3 Speakers Scenario} Using SDR, STOI Metrics.}
   \label{fig:multispeaker}
         \vspace{-3mm}
\end{figure}

\begin{figure}[t]
  \centering
   \includegraphics[width=1.0\linewidth]{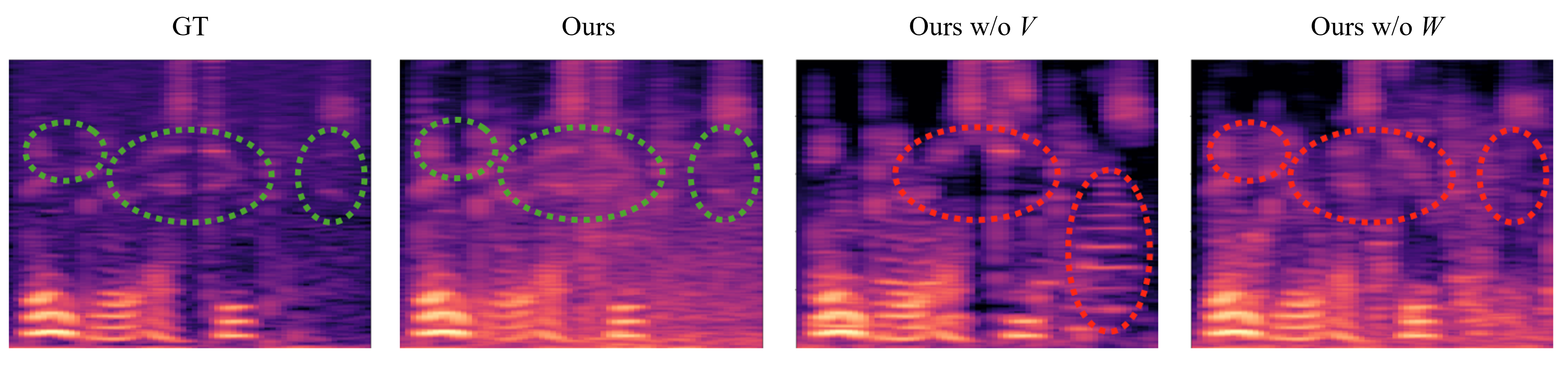}
   \vspace{-7mm}
   \caption{\textbf{\textcolor{black}{Ablation Study} (Mel-Spectrogram Comparison).} Removing $V$ (visual features) leads to noticeable distortions, while the absence of $W$ (Wasserstein distance) results in blurred and less structured reconstructions.}
   \label{fig:melspectrogram_ablation}
   \vspace{-3mm}
\end{figure}

\noindent\textbf{Architecture and Implementation.}
UniVoiceLite adopted an audio-visual Wasserstein autoencoder architecture, where the encoder separately encoded log-magnitude spectrograms and visual features (lip motion and facial attributes) before fusing them to estimate the posterior latent distribution. A visual-conditioned prior was used to predict the latent prior distribution. The decoder reconstructed speech from both posterior and prior samples, guided by a Wasserstein-style loss that combines reconstruction accuracy and latent distribution alignment. The model utilized Tanh for audio activations and ReLU for visual branches. Training was performed using the Adam optimizer with a batch size of $512$, a learning rate of $1e\text{-}4$, and early stopping based on validation loss. The trade-off weight $\lambda$ for the regularization term was set to $0.1$.

\subsection{Experimental Results}
We compared UniVoiceLite with two unsupervised SE models~\cite{RVAE,AV-VAE}, three state-of-the-art SE models~\cite{lu2023mp,cao2022cmgan,kong2020hifi}, and recent SS models, including the audio-only MossFormer2~\cite{zhao2024mossformer2} and audio-visual SS methods~\cite{clearervoice2024,gabbay2017visual,wu2019time}.

\noindent\textbf{Evaluation of Speech Enhancement (SE) Tasks.}
In noisy conditions, UniVoiceLite achieved the highest SDR and STOI across all noise types (Table~\ref{tab:noisy}), despite not being specifically tailored for speech enhancement. While its DNSMOS scores were slightly lower than those of dedicated SE models~\cite{lu2023mp,cao2022cmgan,kong2020hifi}, these models performed poorly on separation tasks (Table~\ref{tab:sota_se}), underscoring UniVoiceLite’s broader applicability. Fig.\ref{fig:melspectrogram} presents qualitative results via mel-spectrogram comparisons, where our model better restores spectral detail.

\noindent\textbf{Evaluation of Speech Separation (SS) Tasks.}
In multi-speaker settings, UniVoiceLite consistently outperformed all baselines in SDR and STOI (Table~\ref{tab:speech_separation}). Although its DNSMOS scores were marginally lower, this metric reflects general perceptual quality and does not explicitly evaluate target speaker preservation. The superior SDR/STOI results demonstrate UniVoiceLite’s effectiveness in extracting the target speaker and suppressing interference. Table~\ref{tab:sota_se} highlights the limited separation ability of SE-focused models, while Table~\ref{tab:sota_ss} shows that UniVoiceLite surpasses both audio-only and audio-visual SS baselines. Fig.\ref{fig:multispeaker} further illustrates UniVoiceLite’s robustness in challenging three-speaker scenarios across various SNR levels, where separation becomes significantly more difficult.

Although some baselines report higher scores in their original papers, our reproduced results are comparatively lower due to a unified and more realistic evaluation setup. Instead of relying on curated benchmarks, we synthesized noisy mixtures using the GRID and DEMAND datasets.
As shown in Fig.\ref{fig:melspectrogram}, the ground truth contains considerable noise, highlighting the difficulty and real-world relevance of our test setting.

\noindent\textbf{Evaluation of Model Efficiency.}
As summarized in Table~\ref{tab:sota_ss}, UniVoiceLite achieves strong performance with only \texttt{2.37M} parameters, significantly fewer than recent audio-visual speech separation models such as VisualVoice~\cite{gabbay2017visual} and AV-ConvTasNet~\cite{wu2019time}. Despite its lightweight design, our model outperforms all baselines in both SDR and STOI. This efficiency is primarily due to three factors: using shallow fully connected layers instead of heavy modules, adopting a low-dimensional latent space, and relying on frozen visual features without end-to-end training.

\begin{table}[t]
\caption{\textbf{\textcolor{black}{Ablation Study} (Importance of Visual Features and Wasserstein Distance)}. $V$ and $W$ are visual features and Wasserstein distance, respectively.}
\label{tab:ablation}
\vspace{1mm}
\centering
\footnotesize{
\setlength{\tabcolsep}{5pt}
\renewcommand{\arraystretch}{1.0}
\begin{tabular}{l|cccc}
\hline
& SDR$\uparrow$  & STOI$\uparrow$  & DNSMOS-s$\uparrow$ & DNSMOS-o$\uparrow$       \\ \hline
Ours w/o W     & -5.80   & 0.28     & 1.69    & 1.43         \\
Ours w/o V    & \multicolumn{1}{c}{10.51}    & 0.85 & 2.98  & 1.93  \\
\textbf{Ours}  & \multicolumn{1}{c}{\textbf{17.82}} & \textbf{0.87} & \textbf{3.11} & \textbf{2.70} \\ \hline
\end{tabular}
}
\vspace{-3mm}
\end{table}

\noindent\textbf{Ablation Study.}
Table~\ref{tab:ablation} evaluates the contribution of key components in our model. Removing visual features leads to a significant degradation in DNSMOS, confirming the importance of lip motion and facial cues as effective visual priors for preserving perceptual clarity. Additionally, replacing the Wasserstein loss with KL divergence results in noticeable drops in SDR and STOI, validating its importance in maintaining a well-structured latent space and avoiding posterior collapse. These findings are further illustrated in Fig.\ref{fig:melspectrogram_ablation}, which visualizes the impact of each component, demonstrating that both visual guidance and Wasserstein regularization are essential for high-quality reconstruction and separation.
\section{Conclusion}
We proposed a lightweight and unsupervised audio-visual Wasserstein autoencoder (UniVoiceLite) that unifies speech enhancement and separation within a single framework. By leveraging visual features and Wasserstein regularization, UniVoiceLite effectively models real-world speech and improves speech intelligibility and stability in noisy, multi-talker environments. Without relying on paired noisy-clean data, the model generalizes well across diverse environments while maintaining speaker independence, offering an efficient and scalable solution for real-world speech processing.
\section*{Acknowledgment}

This work was supported by the Institute of Information \& Communications Technology Planning \& Evaluation (IITP) grant funded by the Korea government (MSIT) (RS-2024-00456709 and RS-2021-II211341, Artificial Intelligence Graduate School Program (Chung-Ang University)).

\bibliographystyle{IEEEtran}
\bibliography{bib}

@string{icassp = "Proc. ICASSP"}

@string{interspeech = "Proc. Interspeech"}

@string{asru = "Proc. ASRU"}

@string{iclr = "Proc. ICLR"}

@string{eccv = "Proc. ECCV"}

@string{cvpr = "Proc. CVPR"}

@article{xu2014regression,
  title={A regression approach to speech enhancement based on deep neural networks},
  author={Xu, Yong and Du, Jun and Dai, Li-Rong and Lee, Chin-Hui},
  journal={IEEE/ACM transactions on audio, speech, and language processing},
  volume={23},
  number={1},
  pages={7--19},
  year={2014}
}

@article{wang2018supervised,
  title={Supervised speech separation based on deep learning: An overview},
  author={Wang, DeLiang and Chen, Jitong},
  journal={IEEE/ACM transactions on audio, speech, and language processing},
  volume={26},
  number={10},
  pages={1702--1726},
  year={2018}
}

@inproceedings{erdogan2015phase,
  title={Phase-sensitive and recognition-boosted speech separation using deep recurrent neural networks},
  author={Erdogan, Hakan and Hershey, John R and Watanabe, Shinji and Le Roux, Jonathan},
  booktitle={ICASSP},
  year={2015}
}

@inproceedings{hershey2016deep,
  title={Deep clustering: Discriminative embeddings for segmentation and separation},
  author={Hershey, John R and Chen, Zhuo and Le Roux, Jonathan and Watanabe, Shinji},
  booktitle={ICASSP},
  year={2016}
}

@article{afouras2018deep,
  title={Deep audio-visual speech recognition},
  author={Afouras, Triantafyllos and Chung, Joon Son and Senior, Andrew and Vinyals, Oriol and Zisserman, Andrew},
  journal={IEEE transactions on pattern analysis and machine intelligence},
  volume={44},
  number={12},
  pages={8717--8727},
  year={2018}
}

@inproceedings{RVAE,
  title={A recurrent variational autoencoder for speech enhancement},
  author={Leglaive, Simon and Alameda-Pineda, Xavier and Girin, Laurent and Horaud, Radu},
  booktitle={ICASSP},
  year={2020}
}

@inproceedings{afouras2018conversation,
  author    = {Afouras, Triantafyllos and Chung, Joon Son and Zisserman, Andrew},
  title     = {The Conversation: Deep Audio-Visual Speech Enhancement},
  booktitle = {Interspeech},
  year      = {2018}
}

@InProceedings{afouras2019my,
  author    = {Afouras, Triantafyllos and Chung, Joon Son and Zisserman, Andrew},
  title     = {My lips are concealed: Audio-visual speech enhancement through obstructions},
  booktitle = {Interspeech},
  year      = {2019}
}

@inproceedings{ephrat2018looking,
  author    = {Ephrat, Ariel and Mosseri, Inbar and Lang, Oran and Dekel, Tali and Wilson, Kevin and Hassidim, Avinatan and Freeman, William T. and Rubinstein, Michael},
  title     = {Looking to Listen at the Cocktail Party: A Speaker-Independent Audio-Visual Model for Speech Separation},
  booktitle = {SIGGRAPH},
  year      = {2018}
}

@article{ideli2019audio,
  title={Audio-visual speech processing using deep learning techniques},
  author={Ideli, Elham},
  year={2019},
  publisher={Simon Fraser University}
}

@inproceedings{inan122019evaluating,
  title={Evaluating audiovisual source separation in the context of video conferencing},
  author={Inan12, Berkay and Cernak, Milos and Grabner23, Helmut and Tukuljac, Helena Peic and Pena, Rodrigo CG and Ricaud, Benjamin},
  booktitle={Interspeech},
  year={2019}
}

@inproceedings{ideli2019visually,
  title={Visually assisted time-domain speech enhancement},
  author={Ideli, Elham and Sharpe, Bruce and Baji{\'c}, Ivan V and Vaughan, Rodney G},
  booktitle={GlobalSIP},
  year={2019}
}

@inproceedings{afouras2020self,
  title={Self-supervised learning of audio-visual objects from video},
  author={Afouras, Triantafyllos and Owens, Andrew and Chung, Joon Son and Zisserman, Andrew},
  booktitle={ECCV},
  year={2020}
}

@inproceedings{gabbay2017visual,
  title={Visual speech enhancement},
  author={Gabbay, Aviv and Shamir, Asaph and Peleg, Shmuel},
  booktitle={Interspeech},
  year={2018}
}

@article{king2009dlib,
  title={Dlib-ml: A machine learning toolkit},
  author={King, Davis E},
  journal={The Journal of Machine Learning Research},
  volume={10},
  pages={1755--1758},
  year={2009}
}

@inproceedings{shi2022learning,
  author    = {Shi, Bowen and Hsu, Wei-Ning and Lakhotia, Kushal and Mohamed, Abdelrahman},
  title     = {Learning Audio-Visual Speech Representation by Masked Multimodal Cluster Prediction},
  booktitle = {Interspeech},
  year      = {2022}
}

@article{cooke2006audio,
  title={An audio-visual corpus for speech perception and automatic speech recognition},
  author={Cooke, Martin and Barker, Jon and Cunningham, Stuart and Shao, Xu},
  journal={The Journal of the Acoustical Society of America},
  volume={120},
  number={5},
  pages={2421--2424},
  year={2006}
}

@article{AV-VAE,
  title={Audio-visual speech enhancement using conditional variational auto-encoders},
  author={Sadeghi, Mostafa and Leglaive, Simon and Alameda-Pineda, Xavier and Girin, Laurent and Horaud, Radu},
  journal={IEEE/ACM Transactions on Audio, Speech, and Language Processing},
  volume={28},
  pages={1788--1800},
  year={2020}
}

@inproceedings{tolstikhin2017wasserstein,
  author    = {Tolstikhin, Ilya and Bousquet, Olivier and Gelly, Sylvain and Sch{\"o}lkopf, Bernhard},
  title     = {Wasserstein Auto-Encoders},
  booktitle = {ICLR},
  year      = {2018}
}

@inproceedings{VAE,
  author    = {Kingma, Diederik P. and Welling, Max},
  title     = {Auto-Encoding Variational Bayes},
  booktitle = {ICLR},
  year      = {2014}
}

@inproceedings{leglaive2018variance,
  title={A variance modeling framework based on variational autoencoders for speech enhancement},
  author={Leglaive, Simon and Girin, Laurent and Horaud, Radu},
  booktitle={MLSP},
  year={2018}
}

@inproceedings{fang2021variational,
  title={Variational autoencoder for speech enhancement with a noise-aware encoder},
  author={Fang, Huajian and Carbajal, Guillaume and Wermter, Stefan and Gerkmann, Timo},
  booktitle={ICASSP},
  year={2021}
}

@inproceedings{reddy2021dnsmos,
  title={DNSMOS: A non-intrusive perceptual objective speech quality metric to evaluate noise suppressors},
  author={Reddy, Chandan KA and Gopal, Vishak and Cutler, Ross},
  booktitle={ICASSP},
  year={2021}
}

@inproceedings{lu2023mp,
  author    = {Lu, Yu-Xuan and Ai, Yuning and Ling, Zhen-Hua},
  title     = {MP-SENet: A Speech Enhancement Model with Parallel Denoising of Magnitude and Phase Spectra},
  booktitle = {Interspeech},
  year      = {2021}
}

@inproceedings{cao2022cmgan,
  author    = {Cao, Ruizhe and Abdulatif, Sherif and Yang, Bin},
  title     = {CMGAN: Conformer-based Metric GAN for Speech Enhancement},
  booktitle = {Interspeech},
  year      = {2022}
}

@inproceedings{kong2020hifi,
  title={Hifi-gan: Generative adversarial networks for efficient and high fidelity speech synthesis},
  author={Kong, Jungil and Kim, Jaehyeon and Bae, Jaekyoung},
  booktitle={NIPS},
  pages={17022--17033},
  year={2020}
}

@article{MCEM,
  title={A Monte Carlo implementation of the EM algorithm and the poor man's data augmentation algorithms},
  author={Wei, Greg CG and Tanner, Martin A},
  journal={Journal of the American statistical Association},
  volume={85},
  number={411},
  pages={699--704},
  year={1990},
  publisher={Taylor \& Francis}
}

@inproceedings{zhao2024mossformer2,
  title={Mossformer2: Combining transformer and rnn-free recurrent network for enhanced time-domain monaural speech separation},
  author={Zhao, Shengkui and Ma, Yukun and Ni, Chongjia and Zhang, Chong and Wang, Hao and Nguyen, Trung Hieu and Zhou, Kun and Yip, Jia Qi and Ng, Dianwen and Ma, Bin},
  booktitle={ICASSP},
  pages={10356--10360},
  year={2024}
}

@misc{clearervoice2024,
  title        = {ClearerVoice-Studio},
  year         = {2024},
  howpublished = {\url{https://github.com/modelscope/ClearerVoice-Studio}},
  note         = {Accessed: 2025-05-05}
}

@inproceedings{thiemann2013diverse,
  title={The diverse environments multi-channel acoustic noise database (demand): A database of multichannel environmental noise recordings},
  author={Thiemann, Joachim and Ito, Nobutaka and Vincent, Emmanuel},
  booktitle={POMA},
  year={2013}
}

@inproceedings{saijo2023single,
  title={A Single Speech Enhancement Model Unifying Dereverberation, Denoising, Speaker Counting, Separation, and Extraction},
  author={Saijo, Kohei and Zhang, Wangyou and Wang, Zhong-Qiu and Watanabe, Shinji and Kobayashi, Tetsunori and Ogawa, Tetsuji},
  booktitle={ASRU},
  pages={1--6},
  year={2023}
}

@inproceedings{wu2019time,
  title={Time domain audio visual speech separation},
  author={Wu, Jian and Xu, Yong and Zhang, Shi-Xiong and Chen, Lian-Wu and Yu, Meng and Xie, Lei and Yu, Dong},
  booktitle={ASRU},
  pages={667--673},
  year={2019}
}

@inproceedings{baevski2020wav2vec,
  title={wav2vec 2.0: A framework for self-supervised learning of speech representations},
  author={Baevski, Alexei and Zhou, Yuhao and Mohamed, Abdelrahman and Auli, Michael},
  booktitle={NIPS},
  pages={12449--12460},
  year={2020}
}

@inproceedings{parkhi2015deep,
  title={Deep face recognition},
  author={Parkhi, Omkar and Vedaldi, Andrea and Zisserman, Andrew},
  booktitle={BMVC},
  year={2015}
}

@inproceedings{schroff2015facenet,
  title={Facenet: A unified embedding for face recognition and clustering},
  author={Schroff, Florian and Kalenichenko, Dmitry and Philbin, James},
  booktitle={CVPR},
  pages={815--823},
  year={2015}
}

@inproceedings{kalkhorani2023time,
  title={Time-domain transformer-based audiovisual speaker separation},
  author={Kalkhorani, V Ahmadi and Kumar, Anurag and Tan, Ke and Xu, Buye and Wang, D},
  booktitle={Interspeech},
  year={2023}
}

@inproceedings{jain2024lstmse,
  author    = {Jain, Arnav and Sanjotra, Jasmer Singh and Choudhary, Harshvardhan and Agrawal, Krish and Shah, Rupal and Jha, Rohan and Sajid, M. and Hussain, Amir and Tanveer, M.},
  title     = {LSTMSE-Net: Long Short Term Speech Enhancement Network for Audio-Visual Speech Enhancement},
  booktitle = {AVSEC},
  year      = {2024}
}

\end{document}